\documentclass[sigconf]{acmart}

\usepackage{booktabs} 
\usepackage{bm}
\usepackage{bbm}
\usepackage{amsmath}
\usepackage{array}
\usepackage{textcomp}
\usepackage{subfigure}

\setcopyright{rightsretained}

\acmDOI{~/~}

\acmISBN{~/~/~}

\acmConference[~]{~}{~}{~} 
\acmYear{~}
\copyrightyear{~}

\begin{document}
\title{Generative Models for Learning from Crowds}

\author{Chi Hong}
\affiliation{%
  \institution{Tsinghua National Laboratory for Information Science and Technology}
  \institution{Department of Computer Science and Technology,Tsinghua University}
  \city{Beijing} 
  \state{China} 
}
\email{hc15@mails.tsinghua.edu.cn}

\begin{abstract}
In this paper, we propose generative probabilistic models for label aggregation. We use Gibbs sampling and a novel variational inference algorithm to perform the posterior inference. Empirical results show that our methods consistently outperform state-of-the-art methods.
\end{abstract}

%
%






\maketitle

\section{Introduction}
Nowadays, huge amounts of data are produced from various sources which allows people to build models for all kinds of applications. Usually, these data cannot be directly used for model building, they should be labeled at first. It is expensive and time-consuming to employ domain experts to manually label data. Recently, crowdsourcing has become a cheap and efficient way to make large training datasets \cite{eydaapproximating,wang2016blind,kamar2012combining,welinder2010multidimensional,li2017reliable,chen2013pairwise,sung2017classification}. Online platforms such as Crowdflower\footnote{http://crowdflower.com/} and Amazon Mechanical Turk\footnote{https://www.mturk.com/mturk} can split large dataset into small parts and distribute these small labeling tasks to workers \cite{whitehill2009whose,li2016crowdsourcing,kazai2016quality,zhuang2015leveraging}. However, these non-professional workers may provide noisy labels. The accuracy of each worker may be lower than expected. To improve the accuracy, it is important to aggregate the noisy labels and infer the true labels.

Note that each item may be labeled multiple times by different workers. The most straightforward way to infer the true label for each item is to simply choose the most frequent label. This approach is called as majority voting. In most cases, it has significantly better performance than single workers \cite{zhou2012learning,snow2008cheap}. However, majority voting implicitly assumes that all workers are equally good and treats each item independently. For a specific case, majority voting will make mistake if there are few true labels given by experienced workers and lots of identical error labels generated by novices. To overcome this problem, some methods take into account the worker reliability in label aggregation. Dawid and Skene \cite{dawid1979maximum} associated each worker with a confusion matrix which can evaluate the reliabilities and potential biases of the workers. For a given worker, each element of the confusion matrix is a probability that the worker labels items in one class as another. 

In recent years, there are many works which also use confusion matrices to build label aggregation models \cite{raykar2010learning,kim2012bayesian,kamar2015identifying,raykar2010learning,kim2012bayesian}. However, in these methods, each worker only uses one confusion matrix to generate labels for items. There is a one-to-one correspondence between an worker and an confusion matrix. They don't consider the characteristics of items, such as difficulty. It is somewhat unreasonable. Actually, different items may have different difficulty levels. For example, there are two face photos of the same person, one is very clear, the other is blurred. A worker is more likely to make mistakes when recognizing the blurred one. He or she should have different confusion in the face of items (face photos) with different difficulties (blurred or clear).

Different from these methods, in this paper, we set a confusion matrix for each worker-difficulty level pair. The confusion matrices is not only correlate to workers, but also correlate to item difficulty levels. We haven't seen any previous work taking this into account. We assume that the collected labels are generated by a distribution over the confusion matrices, item difficulties, true labels and labels. Based on this assumption, we propose models which utilize item difficulty information in the process of label aggregation. In this paper, our main contributions are as follows:

\begin{itemize}
\item{We define a generative probabilistic model that considers item difficulty in label aggregation. Its model parameters can be simply inferred by Gibbs sampling. We call this model as Item Difficulty-Based Label Aggregation (IDBLA) model.}
\item{We propose a variation of the IDBLA model, considering the existence of particularly simple items and particularly difficult items.}
\item{Gibbs sampling is hard to diagnose convergence. So we derive a novel variational inference algorithm for the IDBLA model. The proposed algorithm can converge to a good solution within a few iterations.}
\item{We also design a method to preliminarily predict the true label and difficulty of each item. The predicted results are used to initialize our models. The initialization is important for the performance}
\end{itemize}
We conduct our experiments on three real datasets and one synthetic dataset. The experimental results show that our methods have better performance than state-of-the-art label aggregation methods.

The structure of the paper is as follows. Section 2 introduces the related work. Section 3 introduces the preliminary notation, the majority voting algorithm and the DS-EM algorithm. Section 4 illustrates our methods in detail. Section 5 introduces the experiments and the results. Section 6 concludes the paper.

\section{Related Work}

Raykar et al. \cite{raykar2010learning} proposed the two two-coin model for binary labeling tasks, which can be seen as a variation of the confusion matrix. In their method, the labels generated by workers are directly used for learning. Estimating the reliability of the workers and the true labels of the items is a byproduct of the method. In Minimax Conditional Entropy (MMCE) \cite{zhou2012learning}, each worker-item pair is related to a independent distribution. The authors used a minimax entropy method to estimate the distributions and infer the true labels. Kim and Ghahramani \cite{kim2012bayesian} used the confusion matrix to define their Bayesian Classifier Combination (BCC) model which is a Bayesian extension of Dawid and Skene's method. Venanzi et al. \cite{venanzi2014community} extended the BCC model and proposed the CommunityBCC model, which groups workers into communities. The workers that belong to the same community have the similar confusion matrices. CommunityBCC is particularly suit for sparse dataset which doesn't have enough labels to learn a large number of parameters. Nguyen et al. \cite{nguyen2016correlated} also proposed a graphical model based approach for crowdsourcing. This model can capture the correlations between entries in worker confusion matrices and use the correlations to improve the estimation of the confusion matrices.

There are many other methods besides confusion matrix-based methods. Zhou and He \cite{zhou2016crowdsourcing} proposed a approach which based on tensor augmentation and completion for crowdsourcing. The collected labels are represented as tensor. The GLAD model \cite{whitehill2009whose} is based on parameters which represent the expertise of workers and difficulties of items. The authors used the Expectation-Maximization algorithm as the learning and inference algorithm. GLAD is a label aggregation model for binary labeling tasks. It simultaneously considers the expertise of each worker and the difficulty of each item. Liu et al. \cite{liu2012variational} also used a single parameter to describe the reliability of a worker. Gaunt et al. \cite{gaunt2016training} trained a deep neural network for label aggregation. Unlike Bayesian network based methods, this method doesn't need to make assumptions for the model definition. However, it is not adapt to the case of incomplete data where some workers labeled only few items. Karger et al. \cite{karger2011iterative} put different weights on workers in their model. This work extended the majority voting algorithm.  In recent years, Multi-Armed Bandits based method \cite{tran2014efficient} has been proposed for crowdsourcing. There are also many works predict the truth from a set of conflicting views \cite{galland2010corroborating,yu2014wisdom}.

Markov chain Monte Carlo methods \cite{george1993variable,griffiths2004finding,kim1999state} and variational inference methods \cite{wang2011online,wainwright2008graphical,kurihara2007collapsed,teh2007collapsed,bishop2006pattern,he2015hawkestopic} have been proposed for many models. These works provided us with references. However, it is hard to find a general algorithm, a concrete model needs a concrete analysis.

\section{Preliminary}

\subsection{Notation}

In the aggregation problem, we consider that there are $K$ workers and $I$ items. In each task, the workers are required to label one item. $L_{i,k}$ is the label of item $i$ that labeled by worker $k$, where $L_{i,k} \in \left\{1,...,C\right\}$. $C$ is the number of classes. $\bm{L}$ is a collection of all the collected labels. Note that each worker may only labels a part of the dataset, if worker $k$ hasn't labeled item $i$, then $L_{i,k} = None$. $T_i$ is the true label of item $i$ and $\bm{T} = \{T_1,...,T_I\}$ includes all the true labels.

\subsection{Majority Voting}

Majority voting is a simple and straightforward way to aggregate labels. For each item $i$, it selects the most frequent answer as the true label $T_i$. This can be represented as:
\begin{equation}
T_i = \mathop{\arg\max}_{l\in\left\{1,...,C\right\}}\sum_{k=1}^{K}\mathbbm{I}(L_{i,k}=l)
\end{equation}
where $\mathbbm{I}(\cdot)$ is the indicator function taking the value $1$ when the predicate is true, and 0 otherwise. Majority voting is easy to implement and normally can generate high quality aggregation result although it only considers each item independently. Recent years, there are several works have extended this method \cite{tian2015max,li2014error}.

\subsection{DS-EM}
DS-EM \cite{dawid1979maximum} is a classic generative approach proposed by Dawid and Skene. It assumes that the worker has consistent performances in different labeling tasks and let $\phi^{(k)}_{t,c}$ be the probability of worker $k$ labels an item as class $c$ when the true label of the item is $t$. It also assumes that $T_i$ has a multinomial distribution, $p(T_i=t)=p_t$, where $i \in \{1,...,I\}$ and $t \in \{1,...,C\}$. DS-EM uses an expectation-maximization (EM) algorithm to obtain maximum likelihood estimates of $\bm{\phi}$ and $\bm{T}$.

\textbf{E-step}: Estimating $\bm{T}$ by using:
\begin{equation*}
\begin{split}
p(T_i=t|\bm{L},\bm{\phi}) &\propto p(T_i=t)p(\bm{L}_i|T_i=t,\bm{\phi}) \\
&\propto p_t\prod_{k=1}^{K}\prod_{l=1}^{C}(\phi_{t,l}^{(k)})^{\mathbbm{I}(L_{i,k}=l)}
\end{split}
\end{equation*}
where $\bm{L}_i=\{L_{i,1},...,L_{i,k}\}$. The value of $p_t$ can be estimated by $\hat{p}_t=\sum_{i=1}^{I}\mathbbm{I}(T_i=t)/I$.

\textbf{M-step}: The likelihood for the full data is:
\begin{equation*}
M(\bm{\phi})=\prod_{i=1}^{I}\prod_{t=1}^{C}\left\{p_t\prod_{k=1}^{K}\prod_{l=1}^{C}(\phi_{t,l}^{(k)})^{\mathbbm{I}(L_{i,k}=l)}\right\}^{\mathbbm{I}(T_i=t)}
\end{equation*}
where the values of $\bm{T}$ and $\{p_1,...,p_C\}$ are known, then $\bm{\phi}$ can be estimated by the maximum-likelihood estimation: $\bm{\hat{\phi}} = \mathop{\arg\max}_{\bm{\phi}}M(\bm{\phi})$. Note that this optimisation problem can be analytically solved.

\section{Methods}

As described in Section 3.3, DS-EM associates each worker $k$ with a probabilistic confusion matrix $\bm{\phi}^{(k)}$. Recently, there are many works \cite{zhang2014spectral,liu2012variational,raykar2010learning,kim2012bayesian} use this kind of confusion matrix to evaluate the ability of a worker, and build their own models.

As mentioned in Section 1, these models don't consider the characteristics of the labeling tasks, such as the difficulties of items. In these models, the confusion matrix is only correlate to worker $k$. It is somewhat unreasonable. Therefore, we try predicting and utilizing the item difficulty information in label aggregation. In our models, we use two indexes for the confusion matrix, one is worker $k$ and the other is difficulty level $h$. Next, we will describe our methods detailedly, including model representation, inference and parameter initialization.

\subsection{Item Difficulty-Based Label Aggregation Model}

 In our model, each item $i$ correlates to a difficulty level $Q_i \in \left\{1,...,H\right\}$, where $H$ is the number of difficulty levels. We set an independent probabilistic confusion matrix $\bm{\pi}^{(k,h)}$ for each worker-difficulty level pair $(k,h)$. Consider item $i$ with true label $T_i \in \left\{1,...,C\right\}$ and difficulty level $Q_i$, we let $p(L_{i,k}|T_i,Q_i)=\pi_{T_i,L_{i,k}}^{(k,Q_i)}$. That is to say the label $L_{i,k}$ that generated by worker $k$ for item $i$ has a multinomial distribution with parameters $\bm{\pi}_{T_i}^{(k,Q_i)}$. We assume that $\bm{\pi}_{t}^{(k,h)}\sim Dirichlet(\omega)$. The Dirichlet distribution is the conjugate prior of the multinomial distribution. The true label $T_i$ and difficulty level $Q_i$ of item $i$ have multinomial distributions.
 
 The generative process is\\
 1. Draw all the confusion matrices $\bm{\pi}_{t}^{(k,h)}\sim Dirichlet(\omega)$.\\
 2. Draw the true label proportion $\bm{\alpha} \sim Dirichlet(\gamma_{\alpha})$.\\
 3. Draw the difficulty level proportion $\bm{\beta} \sim Dirichlet(\gamma_{\beta})$.\\
 4. For each item i,\\
 \hspace*{0.5cm} (a) Draw a true label $T_i \sim Multinomial(\bm{\alpha})$.\\
 \hspace*{0.5cm} (b) Draw a difficulty level $Q_i \sim Multinomial(\bm{\beta})$.\\
 \hspace*{0.5cm} (c) For each label $L_{i,k}$,\\
 \hspace*{1cm} $(c_1)$ Draw the label $L_{i,k} \sim Multinomial(\bm{\pi}_{T_i}^{(k,Q_i)})$.
 
Each worker generates her own labels independently. Labels for different items are independent and identically distributed. Let $\bm{\mu}$ represents the hyperparameters $\omega$, $\gamma_{\alpha}$ and $\gamma_{\beta}$. Then the posterior distribution over the model parameters is:
\begin{multline}
p(\bm{T},\bm{Q},\bm{\pi},\bm{\alpha},\bm{\beta}|\bm{L},\bm{\mu}) \propto p(\bm{T}|\bm{\alpha})p(\bm{\alpha}|\gamma_{\alpha}) \\ p(\bm{Q}|\bm{\beta})  p(\bm{\beta}|\gamma_{\beta})p(\bm{L}|\bm{\pi},\bm{T},\bm{Q})p(\bm{\pi}|\omega)
\end{multline}

Now, we introduce several notations that are needed in the following. $\mathcal{N}_{l}(k,h,t,c)$ is the number of times worker $k$ labels $c$ when the item difficulty level is $h$ and the true label is $t$.
$$\mathcal{N}_{l}(k,h,t,c)=\sum_{i=1}^{I}\mathbbm{I}(L_{i,k}=c,T_i=t,Q_i=h)$$
$\mathcal{N}_{t}(c)$ is the number of items with true label $c$. $\mathcal{N}_{q}(h)$ is the number of items with difficulty level $h$. $\mathcal{N}_{t}(c)=\sum_{i=1}^{I}\mathbbm{I}(T_i=c)$ and  $\mathcal{N}_{q}(h)=\sum_{i=1}^{I}\mathbbm{I}(Q_i=h)$.

According to the generative process, equation (2) can be rewritten as:
\begin{multline}
p(\bm{T},\bm{Q},\bm{\pi},\bm{\alpha},\bm{\beta}|\bm{L},\bm{\mu}) \propto  \left\{\prod_{c=1}^{C}\alpha_{c}^{\mathcal{N}_{t}(c)+\gamma_{\alpha}-1}\right\}
\left\{\prod_{h=1}^{H}\beta_{h}^{\mathcal{N}_{q}(h)+\gamma_{\beta}-1}\right\}\\
\left\{\prod_{k=1}^{K}\prod_{h=1}^{H}\prod_{t=1}^{C}\prod_{c=1}^{C}(\pi_{t,c}^{(k,h)})^{\mathcal{N}_{l}(k,h,t,c)+\omega-1}\right\}
\end{multline}


The parameters $\bm{T}$, $\bm{Q}$, $\bm{\pi}$, $\bm{\alpha}$ and $\bm{\beta}$ can be learned with a Gibbs sampler. For each iteration, we update each parameter by sampling from its conditional distribution given the rest parameters. Here are the conditional distributions for Gibbs sampling which are derived from equation (3):
\begin{equation}
p(T_{i}=t|rest) \propto \alpha_{t}\prod_{k=1}^{K}\prod_{c=1}^{C}(\pi_{t,c}^{(k,Q_i)})^{\mathbbm{I}(L_{i,k}=c)}
\end{equation}
\begin{equation}
p(Q_{i}=h|rest) \propto \beta_{h}\prod_{k=1}^{K}\prod_{c=1}^{C}(\pi_{T_i,c}^{(k,h)})^{\mathbbm{I}(L_{i,k}=c)}
\end{equation}
\begin{equation}
p(\bm{\pi}_{t}^{(k,h)}|rest) \propto \prod_{c=1}^{C}(\pi_{t,c}^{(k,h)})^{\mathcal{N}_{l}(k,h,t,c)+\omega-1}
\end{equation}
\begin{equation}
p(\bm{\alpha}|rest) \propto \prod_{c=1}^{C}\alpha_{c}^{\mathcal{N}_{t}(c)+\gamma_{\alpha}-1}
\end{equation}
\begin{equation}
p(\bm{\beta}|rest) \propto \prod_{h=1}^{H}\beta_{h}^{\mathcal{N}_{q}(h)+\gamma_{\beta}-1}
\end{equation}
We see that the posterior distributions of $\bm{\pi}_{t}^{(k,h)}$, $\bm{\alpha}$ and $\bm{\beta}$ take the form of Dirichlet distributions. The new values of $T_i$ and $Q_i$ are generated by multinomial distributions. So model parameters $\bm{T}$, $\bm{Q}$, $\bm{\pi}$, $\bm{\alpha}$ and $\bm{\beta}$ all can be easily sampled.

\subsection{A Variation of the IDBLA Model}
In the dataset, there may be some very easy items and some very difficult items. We assume that every worker labels the easy items with a high correct rate and labels the difficult items with a very low correct rate. Based on this assumption, we propose the Fixed-IDBLA model which is a simple variation of the IDBLA model. 



In Fixed-IDBLA, we set $\bm{\pi}^{(k,H-1)}$ for easy items and set $\bm{\pi}^{(k,H)}$ for difficult items, where $k \in \left\{1,...,K\right\}$. $\bm{\pi}^{(k,H-1)}$ and $\bm{\pi}^{(k,H)}$ are fixed as:
\begin{equation*}
\bm{\pi}^{(k,H-1)} = 
\begin{pmatrix}
1-\nu & \frac{\nu}{C-1} & \dots & \frac{\nu}{C-1}\\
\frac{\nu}{C-1} & 1-\nu & \dots & \frac{\nu}{C-1}\\
\vdots & \vdots & \ddots & \vdots\\
\frac{\nu}{C-1} & \frac{\nu}{C-1} & \dots & 1-\nu
\end{pmatrix}
\end{equation*}
\begin{equation*}
\bm{\pi}^{(k,H)} = 
\begin{pmatrix}
1-\delta & \frac{\delta}{C-1} & \dots & \frac{\delta}{C-1}\\
\frac{\delta}{C-1} & 1-\delta & \dots & \frac{\delta}{C-1}\\
\vdots & \vdots & \ddots & \vdots\\
\frac{\delta}{C-1} & \frac{\delta}{C-1} & \dots & 1-\delta
\end{pmatrix}
\end{equation*}
where $\nu$ and $\delta$ are constants. We assume that $\bm{\pi}_t^{(k,h)} \sim Dirichlet(\psi)$. The distributions of $\bm{L}$, $\bm{T}$, $\bm{\alpha}$, $\bm{Q}$ and $\bm{\beta}$ are the same as described in 4.1. With the above definitions, we have:
\begin{multline}
p(\bm{L}|\bm{\pi},\bm{T},\bm{Q})p(\bm{\pi}|\bm{\psi}) \propto
\left\{\prod_{k=1}^{K}\prod_{h=1}^{H-2}\prod_{t=1}^{C}\prod_{c=1}^{C}(\pi_{t,c}^{(k,h)})^{\mathcal{N}_{l}(k,h,t,c)+\psi-1}\right\}\\
\left\{\prod_{k=1}^{K}\prod_{h=H-1}^{H}\prod_{t=1}^{C}\prod_{c=1}^{C}(\pi_{t,c}^{(k,h)})^{\mathcal{N}_{l}(k,h,t,c)}\right\}\\
\end{multline}
We again use Gibbs sampling to sample all the unknown model parameters. The latent parameters $\bm{T}$, $\bm{Q}$, $\bm{\alpha}$ and $\bm{\beta}$ are still sampled by (4), (5), (7) and (8). $\bm{\pi}_{t}^{(k,h)}$ is sampled by conditional distribution:
\begin{equation}
p(\bm{\pi}_{t}^{(k,h)}|rest) \propto \prod_{c=1}^{C}(\pi_{t,c}^{(k,h)})^{\mathcal{N}_{l}(k,h,t,c)+\psi-1}
\end{equation}
where $h=1,...,H-2$. Note that the posterior distribution of $\bm{\pi}_{t}^{(k,h)}$ is a Dirichlet distribution.

\subsection{Collapsed Variational Inference}
We have introduced the Gibbs sampling algorithm for the IDBLA and Fixed-IDBLA model. However, Gibbs sampling is hard to diagnose convergence. It also needs large numbers of samples to reduce sampling noise. So we would like to use variational inference algorithm to perform the posterior inference and parameter estimation. The most common variational inference algorithms make the mean-field assumption \cite{wainwright2008graphical}, which may be too strict in practice. Model parameters can be strongly dependent in the true posterior, the mean-field assumption ignores this dependence and may leading to inaccurate estimates of the posterior \cite{teh2007collapsed}. 

In this Section, we derive a collapsed variational inference for the IDBLA model. We don't assume independence between $\bm{\pi}$, $\bm{\alpha}$ and $\bm{\beta}$, but we still assume that variables $\bm{T}$ and $\bm{Q}$ are mutually independent. We use the variational distribution
\begin{equation}
q(\bm{T},\bm{Q},\bm{\pi},\bm{\alpha},\bm{\beta}) = q(\bm{\pi},\bm{\alpha},\bm{\beta}|\bm{T},\bm{Q})\prod_{i=1}^{I}q(T_i|\bm{\lambda}_i)q(Q_i|\bm{\rho}_i)
\end{equation}
to approximate the posterior $p(\bm{T},\bm{Q},\bm{\pi},\bm{\alpha},\bm{\beta}|\bm{L},\bm{\mu})$, where $q(\bm{T},\bm{Q})=\prod_{i=1}^{I}q(T_i|\bm{\lambda}_i)q(Q_i|\bm{\rho}_i)$, the distributions $q(T_i|\bm{\lambda}_i)$ and $q(Q_i|\bm{\rho}_i)$ are multinomial distributions. The data log likelihood can be bounded by
\begin{multline*}
\log p(\bm{L}|\bm{\mu}) = KL(q(\bm{T},\bm{Q},\bm{\pi},\bm{\alpha},\bm{\beta})|p(\bm{T},\bm{Q},\bm{\pi},\bm{\alpha},\bm{\beta}|\bm{L},\bm{\mu}))+\\
\mathcal{L}(q(\bm{T},\bm{Q},\bm{\pi},\bm{\alpha},\bm{\beta})) \geq \mathcal{L}(q(\bm{T},\bm{Q},\bm{\pi},\bm{\alpha},\bm{\beta}))
\end{multline*}
where $KL(\cdot)$ is the Kullback-Leibler divergence, $\mathcal{L}(q(\bm{T},\bm{Q},\bm{\pi},\bm{\alpha},\bm{\beta}))$ is called evidence lower bound (ELBO).
\begin{multline}
\mathcal{L}(q(\bm{T},\bm{Q},\bm{\pi},\bm{\alpha},\bm{\beta})) =\\ \mathbbm{E}_{q(\bm{\pi},\bm{\alpha},\bm{\beta}|\bm{T},\bm{Q})q(\bm{T},\bm{Q})}[\log p(\bm{T},\bm{Q},\bm{\pi},\bm{\alpha},\bm{\beta},\bm{L}|\bm{\mu})] -\\
\mathbbm{E}_{q(\bm{\pi},\bm{\alpha},\bm{\beta}|\bm{T},\bm{Q})q(\bm{T},\bm{Q})}[\log q(\bm{\pi},\bm{\alpha},\bm{\beta}|\bm{T},\bm{Q})q(\bm{T},\bm{Q})] =\\
\iint q(\bm{\pi},\bm{\alpha},\bm{\beta}|\bm{T},\bm{Q})q(\bm{T},\bm{Q}) \log \frac{p(\bm{T},\bm{Q},\bm{\pi},\bm{\alpha},\bm{\beta},\bm{L}|\bm{\mu})}{q(\bm{\pi},\bm{\alpha},\bm{\beta}|\bm{T},\bm{Q})q(\bm{T},\bm{Q})}\,d_{\bm{\pi},\bm{\alpha},\bm{\beta}}\,d_{\bm{T},\bm{Q}}
\end{multline}
where $\mathbbm{E}$ means expectation. We maximize the ELBO with respect to $q(\bm{\pi},\bm{\alpha},\bm{\beta}|\bm{T},\bm{Q})$. The maximum is achieved at $q(\bm{\pi},\bm{\alpha},\bm{\beta}|\bm{T},\bm{Q})=p(\bm{\pi},\bm{\alpha},\bm{\beta}|\bm{L},\bm{T},\bm{Q},\bm{\mu})$ that means the approximation distribution equals the true distribution. Plugging this equation into equation (12), we get
\begin{multline}
\mathcal{L}(q(\bm{T},\bm{Q})) = \mathop{\max}_{q(\bm{\pi},\bm{\alpha},\bm{\beta}|\bm{T},\bm{Q})}\mathcal{L}(q(\bm{T},\bm{Q},\bm{\pi},\bm{\alpha},\bm{\beta})) = \\
\mathbbm{E}_{q(\bm{T},\bm{Q})}[\log p(\bm{T},\bm{Q},\bm{L}|\bm{\mu})] - \mathbbm{E}_{q(\bm{T},\bm{Q})}[\log q(\bm{T},\bm{Q})]
\end{multline}
where we have used $\int q(\bm{\pi},\bm{\alpha},\bm{\beta}|\bm{T},\bm{Q})d_{\bm{\pi},\bm{\alpha},\bm{\beta}}=1$. Marginalizing out $\bm{\pi}$,$\bm{\alpha}$ and $\bm{\beta}$ from the posterior, we have
\begin{multline}
p(\bm{T},\bm{Q}|\bm{L},\bm{\mu}) = \iiint p(\bm{T},\bm{Q},\bm{\pi},\bm{\alpha},\bm{\beta}|\bm{L},\bm{\mu})\,d_{\bm{\pi}}\,d_{\bm{\alpha}}\,d_{\bm{\beta}}\\
\propto \frac{\prod_{c=1}^{C}\Gamma(\mathcal{N}_{t}(c)+\gamma_\alpha)}{\Gamma(I+C\gamma_\alpha)}
\frac{\prod_{h=1}^{H}\Gamma(\mathcal{N}_{q}(h)+\gamma_\beta)}{\Gamma(I+H\gamma_\beta)}\\
\prod_{k=1}^{K}\prod_{h=1}^{H}\prod_{t=1}^{C} \frac{\prod_{c=1}^{C}\Gamma(\mathcal{N}_{l}(k,h,t,c)+\omega)}{\Gamma(\mathcal{N}_{l}(k,h,t,\cdot)+C\omega)}
\end{multline}
where $\mathcal{N}_{l}(k,h,t,\cdot)=\sum_{c=1}^{C}\mathcal{N}_{l}(k,h,t,c)$, $\Gamma(\cdot)$ is the gamma function. According to equation (14), we derive:
\begin{equation}
p(T_i=t|rest) \propto (\mathcal{N}_{t}^{\lnot i}(t)+\gamma_\alpha)\prod_{k \in S_i}\frac{\mathcal{N}_{l}^{\lnot i}(k,Q_i,t,L_{i,k})+\omega}{\mathcal{N}_{l}^{\lnot i}(k,Q_i,t,\cdot)+C\omega}
\end{equation}
\begin{equation}
p(Q_i=h|rest) \propto (\mathcal{N}_{q}^{\lnot i}(h)+\gamma_\beta)\prod_{k \in S_i}\frac{\mathcal{N}_{l}^{\lnot i}(k,h,T_i,L_{i,k})+\omega}{\mathcal{N}_{l}^{\lnot i}(k,h,T_i,\cdot)+C\omega}
\end{equation}
where "$\lnot i$" means that the frequency is calculated over items except item $i$, $S_i$ is a set of workers who have labeled the item $i$. 

Then we can further maximize $\mathcal{L}(q(\bm{T},\bm{Q}))$ with respect to $q(\bm{T},\bm{Q})$ which is factorized. The following maximization process is just like the variational Bayes algorithm \cite{bishop2006pattern}. We optimize the variational parameters $\bm{\lambda}$ and $\bm{\rho}$ to maximize $\mathcal{L}(q(\bm{T},\bm{Q}))$ by the coordinate ascent algorithm. $\bm{\lambda}$ and $\bm{\rho}$ are updated iteratively. For brevity, the derivations of equations (17) and (18) are shown in Appendix A.

\begin{equation}
\lambda_{i,t} = q(T_i=t) \propto \exp(\mathbbm{E}_{q(\bm{T}^{\lnot i},\bm{Q})}[\log p(T_i=t|rest)])
\end{equation}

\begin{equation}
\rho_{i,h} = q(Q_i=h) \propto \exp(\mathbbm{E}_{q(\bm{T},\bm{Q}^{\lnot i})}[\log p(Q_i=h|rest)])
\end{equation}
$\bm{T}^{\lnot i}$ means excluding $T_i$ and $\bm{Q}^{\lnot i}$ means excluding $Q_i$. 

Next, we introduce how to compute $\rho_{i,h}$. Note that the computation of $\lambda_{i,t}$ and $\rho_{i,h}$ are similar. Plugging equation (16) into equation (18), we get
\begin{multline}
\rho_{i,h} \propto \exp \Big(\mathbbm{E}_{q(\bm{T},\bm{Q}^{\lnot i})}\Big[\log (\mathcal{N}_{q}^{\lnot i}(h)+\gamma_\beta) + \\ \sum_{k \in S_i}\big(\log(\mathcal{N}_{l}^{\lnot i}(k,h,T_i,L_{i,k})+\omega)-\log(\mathcal{N}_{l}^{\lnot i}(k,h,T_i,\cdot)+C\omega)\big) \Big]\Big)
\end{multline}
We use the Gaussian approximation to calculate the expectation terms $\mathbbm{E}_{q(\bm{T},\bm{Q}^{\lnot i})}[\log(\mathcal{N}_{l}^{\lnot i}(k,h,T_i,L_{i,k})+\omega)]$, $\mathbbm{E}_{q(\bm{T},\bm{Q}^{\lnot i})}[\log (\mathcal{N}_{q}^{\lnot i}(h)+\gamma_\beta)]$ and $\mathbbm{E}_{q(\bm{T},\bm{Q}^{\lnot i})}[\log(\mathcal{N}_{l}^{\lnot i}(k,h,T_i,\cdot)+C\omega)]$. For brevity, in the following we abbreviate $q(\bm{T},\bm{Q}^{\lnot i})$ as $q^{\lnot i}$. 

Firstly, we apply Gaussian approximation to the expectation $\mathbbm{E}_{q(\bm{T},\bm{Q}^{\lnot i})}[\log(\mathcal{N}_{l}^{\lnot i}(k,h,T_i,L_{i,k})+\omega)]$. Note that the variable
$$\mathcal{N}_{l}^{\lnot i}(k,h,T_i,L_{i,k})=\sum_{j \neq i,L_{j,k}=L_{i,k}}\mathbbm{I}(Q_j = h,T_j = T_i)$$ is a sum of many independent Bernoulli variables $\mathbbm{I}(Q_j = h,T_j = T_i)$ each with mean $(\bm{\lambda}_j^{\rm T}\bm{\lambda}_i)\rho_{j,h}$ and variance $(\bm{\lambda}_j^{\rm T}\bm{\lambda}_i)\rho_{j,h}\big(1-(\bm{\lambda}_j^{\rm T}\bm{\lambda}_i)\rho_{j,h}\big)$. So $\mathcal{N}_{l}^{\lnot i}(k,h,T_i,L_{i,k})$ can be seen as a Gaussian variable, its mean and variance are:
\begin{equation*}
\mathbbm{E}_{q^{\lnot i}}[\mathcal{N}_{l}^{\lnot i}(k,h,T_i,L_{i,k})] = \sum_{j \neq i,L_{j,k}=L_{i,k}}(\bm{\lambda}_j^{\rm T}\bm{\lambda}_i)\rho_{j,h}
\end{equation*}
\begin{equation*}
\mathrm{Var}_{q^{\lnot i}}[\mathcal{N}_{l}^{\lnot i}(k,h,T_i,L_{i,k})] = \sum_{j \neq i,L_{j,k}=L_{i,k}}(\bm{\lambda}_j^{\rm T}\bm{\lambda}_i)\rho_{j,h}\big(1-(\bm{\lambda}_j^{\rm T}\bm{\lambda}_i)\rho_{j,h}\big)
\end{equation*}
The function $\log(\mathcal{N}_{l}^{\lnot i}(k,h,T_i,L_{i,k})+\omega)$ is approximated by its second-order Taylor expansion about $\mathbbm{E}_{q^{\lnot i}}[\mathcal{N}_{l}^{\lnot i}(k,h,T_i,L_{i,k})]$.
\begin{multline}
\log(\mathcal{N}_{l}^{\lnot i}(k,h,T_i,L_{i,k})+\omega) \approx \log (\mathbbm{E}_{q^{\lnot i}}[\mathcal{N}_{l}^{\lnot i}(k,h,T_i,L_{i,k})]+\omega) +\\
\frac{\mathcal{N}_{l}^{\lnot i}(k,h,T_i,L_{i,k})-\mathbbm{E}_{q^{\lnot i}}[\mathcal{N}_{l}^{\lnot i}(k,h,T_i,L_{i,k})]}{\mathbbm{E}_{q^{\lnot i}}[\mathcal{N}_{l}^{\lnot i}(k,h,T_i,L_{i,k})]+\omega} -\\
\frac{(\mathcal{N}_{l}^{\lnot i}(k,h,T_i,L_{i,k})-\mathbbm{E}_{q^{\lnot i}}[\mathcal{N}_{l}^{\lnot i}(k,h,T_i,L_{i,k})])^{2}}{2(\mathbbm{E}_{q^{\lnot i}}[\mathcal{N}_{l}^{\lnot i}(k,h,T_i,L_{i,k})]+\omega)^{2}}
\end{multline}
Using equation (20), we have
\begin{multline}
\mathbbm{E}_{q^{\lnot i}}[\log(\mathcal{N}_{l}^{\lnot i}(k,h,T_i,L_{i,k})+\omega)] \approx \\ \log (\mathbbm{E}_{q^{\lnot i}}[\mathcal{N}_{l}^{\lnot i}(k,h,T_i,L_{i,k})]+\omega) -
\frac{\mathrm{Var}_{q^{\lnot i}}[\mathcal{N}_{l}^{\lnot i}(k,h,T_i,L_{i,k})]}{2(\mathbbm{E}_{q^{\lnot i}}[\mathcal{N}_{l}^{\lnot i}(k,h,T_i,L_{i,k})]+\omega)^{2}}
\end{multline}

$\mathcal{N}_{q}^{\lnot i}(h)=\sum_{j \neq i}\mathbbm{I}(Q_j = h)$ is a sum of many independent Bernoulli variables $\mathbbm{I}(Q_j = h)$ each with mean $\rho_{j,h}$ and variance $\rho_{j,h}(1-\rho_{j,h})$. 
$\mathcal{N}_{l}^{\lnot i}(k,h,T_i,\cdot)=\sum_{j \neq i,L_{j,k} \neq None}\mathbbm{I}(Q_j = h,T_j = T_i)$ is a sum of many independent Bernoulli variables $\mathbbm{I}(Q_j = h,T_j = T_i)$. The Gaussian approximations applied to $\mathbbm{E}_{q(\bm{T},\bm{Q}^{\lnot i})}[\log (\mathcal{N}_{q}^{\lnot i}(h)+\gamma_\beta)]$ and $\mathbbm{E}_{q(\bm{T},\bm{Q}^{\lnot i})}[\log(\mathcal{N}_{l}^{\lnot i}(k,h,T_i,\cdot)+C\omega)]$ are similarly computed as above. According to (19) and (21) we have the update equation:
\begin{multline}
\rho_{i,h} \propto \exp \Bigg( -\frac{\mathrm{Var}_{q^{\lnot i}}[\mathcal{N}_{q}^{\lnot i}(h)]}{2(\mathbbm{E}_{q^{\lnot i}}[\mathcal{N}_{q}^{\lnot i}(h)]+\gamma_\beta)^{2}} 
+ \\ \sum_{k \in S_i} \Big( -\frac{\mathrm{Var}_{q^{\lnot i}}[\mathcal{N}_{l}^{\lnot i}(k,h,T_i,L_{i,k})]}{2(\mathbbm{E}_{q^{\lnot i}}[\mathcal{N}_{l}^{\lnot i}(k,h,T_i,L_{i,k})]+\omega)^{2}}
+ \\ \frac{\mathrm{Var}_{q^{\lnot i}}[\mathcal{N}_{l}^{\lnot i}(k,h,T_i,\cdot)]}{2(\mathbbm{E}_{q^{\lnot i}}[\mathcal{N}_{l}^{\lnot i}(k,h,T_i,\cdot)]+C\omega)^{2}}\Big)  \Bigg)
(\mathbbm{E}_{q^{\lnot i}}[\mathcal{N}_{q}^{\lnot i}(h)]+\gamma_\beta)\\ 
\prod_{k \in S_i}(\mathbbm{E}_{q^{\lnot i}}[\mathcal{N}_{l}^{\lnot i}(k,h,T_i,L_{i,k})]+\omega)(\mathbbm{E}_{q^{\lnot i}}[\mathcal{N}_{l}^{\lnot i}(k,h,T_i,\cdot)]+C\omega)^{-1}
\end{multline}
where
\begin{equation*}
\mathbbm{E}_{q^{\lnot i}}[\mathcal{N}_{l}^{\lnot i}(k,h,T_i,\cdot)] = \sum_{j \neq i,L_{j,k} \neq None}(\bm{\lambda}_j^{\rm T}\bm{\lambda}_i)\rho_{j,h}
\end{equation*}
\begin{equation*}
\mathrm{Var}_{q^{\lnot i}}[\mathcal{N}_{l}^{\lnot i}(k,h,T_i,\cdot)] = \sum_{j \neq i,L_{j,k} \neq None}(\bm{\lambda}_j^{\rm T}\bm{\lambda}_i)\rho_{j,h}\big(1-(\bm{\lambda}_j^{\rm T}\bm{\lambda}_i)\rho_{j,h}\big)
\end{equation*}
\begin{equation*}
\mathbbm{E}_{q^{\lnot i}}[\mathcal{N}_{q}^{\lnot i}(h)] = \sum_{j \neq i}\rho_{j,h}
\hspace{1em}
\mathrm{Var}_{q^{\lnot i}}[\mathcal{N}_{q}^{\lnot i}(h)] = \sum_{j \neq i}\rho_{j,h}(1-\rho_{j,h}).
\end{equation*}

For completeness, we also show the update equation of $\lambda_{i,t}$. We abbreviate $q(\bm{T}^{\lnot i},\bm{Q})$ as $\hat{q}^{\lnot i}$. 
\begin{multline}
\lambda_{i,t} \propto \exp \Bigg( -\frac{\mathrm{Var}_{\hat{q}^{\lnot i}}[\mathcal{N}_{t}^{\lnot i}(t)]}{2(\mathbbm{E}_{\hat{q}^{\lnot i}}[\mathcal{N}_{t}^{\lnot i}(t)]+\gamma_\alpha)^{2}} 
+ \\ \sum_{k \in S_i} \Big( -\frac{\mathrm{Var}_{\hat{q}^{\lnot i}}[\mathcal{N}_{l}^{\lnot i}(k,Q_i,t,L_{i,k})]}{2(\mathbbm{E}_{\hat{q}^{\lnot i}}[\mathcal{N}_{l}^{\lnot i}(k,Q_i,t,L_{i,k})]+\omega)^{2}}
+ \\ \frac{\mathrm{Var}_{\hat{q}^{\lnot i}}[\mathcal{N}_{l}^{\lnot i}(k,Q_i,t,\cdot)]}{2(\mathbbm{E}_{\hat{q}^{\lnot i}}[\mathcal{N}_{l}^{\lnot i}(k,Q_i,t,\cdot)]+C\omega)^{2}}\Big)  \Bigg)
(\mathbbm{E}_{\hat{q}^{\lnot i}}[\mathcal{N}_{t}^{\lnot i}(t)]+\gamma_\alpha)\\ 
\prod_{k \in S_i}(\mathbbm{E}_{\hat{q}^{\lnot i}}[\mathcal{N}_{l}^{\lnot i}(k,Q_i,t,L_{i,k})]+\omega)(\mathbbm{E}_{\hat{q}^{\lnot i}}[\mathcal{N}_{l}^{\lnot i}(k,Q_i,t,\cdot)]+C\omega)^{-1}
\end{multline}
where
\begin{equation*}
\mathbbm{E}_{\hat{q}^{\lnot i}}[\mathcal{N}_{l}^{\lnot i}(k,Q_i,t,L_{i,k})] = \sum_{j \neq i,L_{j,k}=L_{i,k}}(\bm{\rho}_j^{\rm T}\bm{\rho}_i)\lambda_{j,t}
\end{equation*}
\begin{equation*}
\mathrm{Var}_{\hat{q}^{\lnot i}}[\mathcal{N}_{l}^{\lnot i}(k,Q_i,t,L_{i,k})] = \sum_{j \neq i,L_{j,k}=L_{i,k}}(\bm{\rho}_j^{\rm T}\bm{\rho}_i)\lambda_{j,t}\big(1-(\bm{\rho}_j^{\rm T}\bm{\rho}_i)\lambda_{j,t}\big)
\end{equation*}
\begin{equation*}
\mathbbm{E}_{\hat{q}^{\lnot i}}[\mathcal{N}_{l}^{\lnot i}(k,Q_i,t,\cdot)] = \sum_{j \neq i,L_{j,k} \neq None}(\bm{\rho}_j^{\rm T}\bm{\rho}_i)\lambda_{j,t}
\end{equation*}
\begin{equation*}
\mathrm{Var}_{\hat{q}^{\lnot i}}[\mathcal{N}_{l}^{\lnot i}(k,Q_i,t,\cdot)] = \sum_{j \neq i,L_{j,k} \neq None}(\bm{\rho}_j^{\rm T}\bm{\rho}_i)\lambda_{j,t}\big(1-(\bm{\rho}_j^{\rm T}\bm{\rho}_i)\lambda_{j,t}\big)
\end{equation*}
\begin{equation*}
\mathbbm{E}_{\hat{q}^{\lnot i}}[\mathcal{N}_{t}^{\lnot i}(t)] = \sum_{j \neq i}\lambda_{j,t}
\hspace{1em}
\mathrm{Var}_{\hat{q}^{\lnot i}}[\mathcal{N}_{t}^{\lnot i}(t)] = \sum_{j \neq i}\lambda_{j,t}(1-\lambda_{j,t}).
\end{equation*}

\subsection{Parameter Initialization}
In many crowdsourcing methods, such as BCC \cite{kim2012bayesian}, DS-EM \cite{dawid1979maximum}, CrowdSVM \cite{tian2015max}, and DiagCov \cite{nguyen2016correlated}, the unknown true labels are initialized by majority voting to avoid bad local optima. 

In our models, we have introduced the concept of item difficulty. Both parameters $T$ and $Q$ need to be initialized. We manage to design a low overhead method to initialize them. Note that instead of making a precise prediction, we just make a preliminarily prediction of true labels $T$ and difficulty levels $Q$ to avoid bad local optima in the non-convex optimization.

In actually, the ground truth labels are unknown. So we use majority voting to initialize $\bm{T}$, and approximate $\bm{T}$ as a collection of the ground truth labels. Based on $\bm{T}$, we calculate the correct rate $R_k$ for each worker $k$:
\begin{equation*}
R_k = \frac{\sum_{i=1}^{I}\mathbbm{I}(L_{i,k}=T_i)}{\sum_{i=1}^{I}\mathbbm{I}(L_{i,k}\neq None)}
\end{equation*}
Then, let $\lambda_k = xR_k$ be the ability of worker $k$, where $x$ is a constant. We assume that:
\begin{equation}
p(L_{i,k}=T_i|\lambda_k, \epsilon_i) = \frac{1}{1+(C-1)e^{-\lambda_k\epsilon_i}},
\end{equation}
where $1/\epsilon_i \in [0,+\infty)$ is the difficulty of item $i$. 

We see that as the item difficulty $1/\epsilon_i$ increases, the probability of labeling the item correctly decreases toward $1/C$. That means for the most difficult item the worker just arbitrarily chooses a label. When $1/\epsilon_i$ decreases toward $0$, the probability of labeling the item correctly increases toward $1$. Using (24), we have the likelihood of the observed labels:
\begin{equation}
p(\bm{L}|\bm{T},\bm{\lambda},\bm{\epsilon})=\prod_{i=1}^{I}\prod_{k \in S_i}p(L_{i,k}|T_i,\lambda_k, \epsilon_i)
\end{equation}
where $\bm{\lambda}=\{\lambda_1,...,\lambda_K\}$, $\bm{\epsilon}=\{\epsilon_1,...,\epsilon_I\}$. We use gradient ascent to locally maximize the log-likelihood $F(\bm{\epsilon})=\ln p(\bm{L}|\bm{T},\bm{\lambda},\bm{\epsilon})$ and get the corresponding value of $\bm{\epsilon}$. Further, according to $\bm{\epsilon}$, $\bm{Q}$ is initialized by dividing the item into $H$ groups which belong to different difficulty levels. 


\section{Experiments}
IDBLA and Fixed-IDBLA are compared with three state-of-the-art algorithms: majority voting, DS-EM and BCC\cite{kim2012bayesian}. We use three real crowdsourcing datasets and one synthetic dataset in our experiments. In the following, we will introduce our experiments in detail, and evaluate the effectiveness of our models.

\subsection{Datasets}

The four datasets are shown in Table~\ref{datasets}. They have different sizes and features. We introduce them in the following sections.

\begin{table}[h]
\caption{Datasets Overview}
\label{datasets}
\begin{center}
\begin{tabular}{c|c|c|c|c}\hline\hline
Dataset & Workers & Items & Labels & Classes \\ \hline
Heartdisease & 12 & 237 & 952 & 2 \\ 
Web Search & 76 & 2653 & 14638 & 5 \\ 
RTE & 164 & 800 & 8000 & 2 \\
Synthetic & 100 & 1000 & 11615 & 5 \\ \hline
\end{tabular}
\end{center}
\end{table}

\subsubsection{Heart Disease Diagnosis}
We got the heart disease instances \cite{Heart1988} and the corresponding ground truth labels from the UC Irvine machine learning repository website. Each instance include attributes such as age, sex, maximum heart rate achieved, etc. We removed the ground truth labels and requested $12$ medical students to label the instances in their spare time. These students volunteer to offer the labels without pay. They have different expertise levels about heart disease diagnosis. For each instance we don't care about the type of heart disease, we only consider whether the patient has heart disease. Each student doesn't have to label all the instances. The number of the instances for heart disease is $87$ and there are other $150$ instances for health. There are $237$ instances in total. We finally got $952$ labels. The average accuracy of the students is $68.59\%$. The student who labeled the most labeled $213$ instances with accuracy of $87.79\%$. Each student labeled at least 43 instances. For the sake of simplicity, in the following we refer to this data set as \textbf{Heartdisease}.

\subsubsection{Web Search Relevance Judgment}
\begin{figure}[h]
\begin{center}
\includegraphics[scale=0.35]{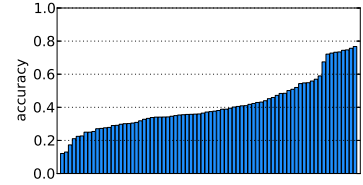}
\end{center}
\caption{Accuracies of workers on the web dateset}
\label{web}
\end{figure}

The \textbf{Web Search} \cite{zhou2012learning} dataset is about web search relevance judgment. In the original dataset, workers are asked to rate a set of 2665 query-URL pairs on a relevance rating scale from 1 to 5. It is difficult to evaluate the worker who only labeled few items. Therefore, we removed the workers who labeled less than 30 items. We also removed the items that haven't ground truth labels. Then we got $2653$ items and $14638$ labels which are offered by $76$ workers. The average accuracy of these workers is $41.08\%$. The accuracy of the best worker is $76.73\%$ and this worker generated 1225 labels. All the labels are offered by workers in MTurk website. The accuracies of the workers are shown in Figure~\ref{web}, each bar represents the accuracy of one worker.

\subsubsection{RTE Dataset}
In this dataset, the tasks are about recognizing textual entailment. \textbf{RTE} \cite{snow2008cheap} contains $8000$ binary labels for 800 documents. The numbers of the documents for each class are all equal to $400$. A total of $164$ labelers participated in the labeling task. The average accuracy of the labelers is $83.70\%$. The labeler who labeled the most generated 800 labels and had a accuracy of $50.63\%$.

\subsubsection{Synthetic Dataset}
In addition to the above three real data sets, we also made a \textbf{Synthetic} dataset. We generated 1000 items and 100 workers. The true label of each item is sampled from $\{1,2,3,4,5\}$ with probabilities $\{0.18,0.27,0.45,0.05,0.05\}$, the classes are imbalanced. In order to simulate the real situation, each worker $k$ labels a item with probability $\rho_k$. The maximum value in $\bm{\rho}=\{\rho_1,...,\rho_{100}\}$ is $0.74$, most elements in $\bm{\rho}$ get values in the interval $[0.03, 0.2]$. The average accuracy of the workers is $41.92\%$. The best worker has a accuracy of $83.49\%$. Finally, we got a total of $11615$ labels.

\begin{table*}[ht]
\caption{Error-rates (\%) of Methods}
\label{error-rates}
\begin{center}
\begin{tabular}
{p{3cm}<{\centering}|p{2cm}<{\centering}p{2cm}<{\centering}p{2cm}<{\centering}p{2cm}<{\centering}}\hline\hline
Method & Heartdisease & Web Search & RTE & Synthetic \\ \hline
Majority Voting & 24.89 & 24.76 & 9.88 & 19.80 \\ 
DS-EM & 18.99 & 18.58 & 7.50 & 13.50 \\
BCC & 18.82 & 21.57 & 7.15 & 11.26 \\ \hline\hline
IDBLA & 16.03 & \textbf{16.66} & \textbf{7.13} & \textbf{9.50} \\ 
Fixed-IDBLA & \textbf{15.61} & 18.77 & 7.50 & \textbf{9.50} \\ \hline
\end{tabular}
\end{center}
\end{table*}

\subsection{Setups}
Majority voting and DS-EM are implemented according to the introduction in Section 3. In a specific task, if there are multiple most frequent classes for the item, the majority voting algorithm will randomly choose one among them. 

In order to avoid bad local optima, DS-EM and BCC use majority voting to initialize the unknown true labels. For BCC, the hyperparameters are set as described in Kim's paper\cite{kim2012bayesian}. The model parameters are also initialized according to the paper's description.

For IDBLA model, $\omega$, $\gamma_{\alpha}$ and $\gamma_{\beta}$ are all set as $1.0$. Which means the distributions of $\bm{\pi}_{t}^{(k,h)}$, $\bm{\alpha}$ and $\bm{\beta}$ have uninformed priors. $\bm{T}$ and $\bm{Q}$ are are initialized as described in Section 4.4.$\bm{\alpha}$ is initialized by the result of counting $\bm{T}$, 
$
\alpha_{c} = \frac{\sum_{i=1}^{I}\mathbbm{I}(T_i=c)}{I}. 
$
$\bm{\beta}$ is initialized by the result of counting $\bm{Q}$, 
$
\beta_{h} = \frac{\sum_{i=1}^{I}\mathbbm{I}(Q_i=h)}{I}. 
$
According to $\bm{T}$, $\bm{Q}$ and the known $\bm{L}$, $\pi_{t,c}^{(k,h)}$ can be initialized by:
\begin{equation*}
\pi_{t,c}^{(k,h)} = \frac{\sum_{i=1}^{I}\mathbbm{I}(L_{i,k}=c,T_i=t,Q_i=h)}{\sum_{i=1}^{I}\mathbbm{I}(L_{i,k} \neq None,T_i=t,Q_i=h)}
\end{equation*}
For Fixed-IDBLA model, $\nu$ is set as $0.1$, $\delta$ is set as $0.8$. The other model parameters and hyperparameters are set the same as in IDBLA.

\textbf{$\bm{H}$ selection}:For IDBLA and Fixed-IDBLA models, the number of difficulty levels $H$ should be determined. Given a value $\hat{H}$ and a dataset, the corresponding likelihood $p(\bm{L}|\hat{\bm{\pi}},\hat{\bm{T}},\hat{\bm{Q}})$ could be computed. So, for each dataset, we can try several values of $H$ and select the value which generates the maximum likelihood. Note that the likelihood $p(\bm{L}|\hat{\bm{\pi}},\hat{\bm{T}},\hat{\bm{Q}})$ and the accuracy are not directly related.

\subsection{Experimental Results}

We used a PC with Intel Core i5 2.6GHz CPU and 8GB RAM for our experiments. In order to guarantee the fairness of the experiments. We use uninformed priors in all methods. We use majority voting to initialize parameters in all methods. Every method have the same input data format.

\subsubsection{Error-rates of Methods}

Firstly, we use error rate to evaluate the performance of each method. For BCC, IDBLA and Fixed-IDBLA, we sampled 500 samples in each run. On each dataset, we executed every method independently 10 times and averaged the error rates.

The error rates of the methods are shown in Table~\ref{error-rates}. Compared with Figure~\ref{web}, we can see that these label aggregation methods generate results obviously better than single workers. The best result for each dataset is highlighted in bold. In our experiments, IDBLA consistently outperforms majority voting, DS-EM and BCC across all datasets. On the Heartdisease dataset, Fixed-IDBLA has the best performance. IDBLA achieves a lower error rate than Fixed-IDBLA on the Web Search dataset and RTE dataset. On the Synthetic dataset, IDBLA and Fixed-IDBLA have the same error rate. DS-EM, BCC, IDBLA and Fixed-IDBLA are substantially better than majority voting. They are methods based on confusion matrix. The experimental results show that our methods outperform the baselines.

\subsubsection{Collapsed Variational Inference Iteration Process}

\begin{figure}[h]
\centering
\subfigure[Heartdisease]{
\label{iter:a}
\includegraphics[scale=0.35]{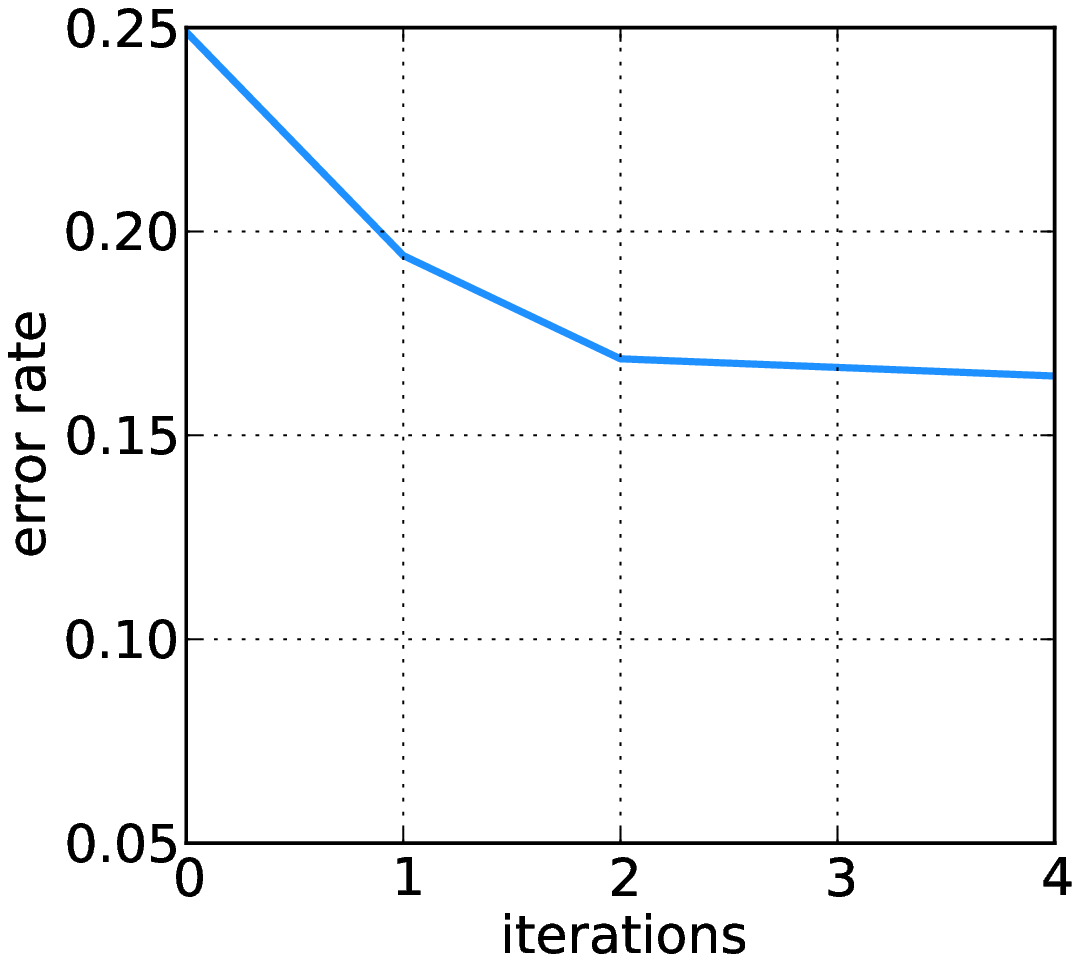}
}
\subfigure[Web Search]{
\label{iter:b}
\includegraphics[scale=0.35]{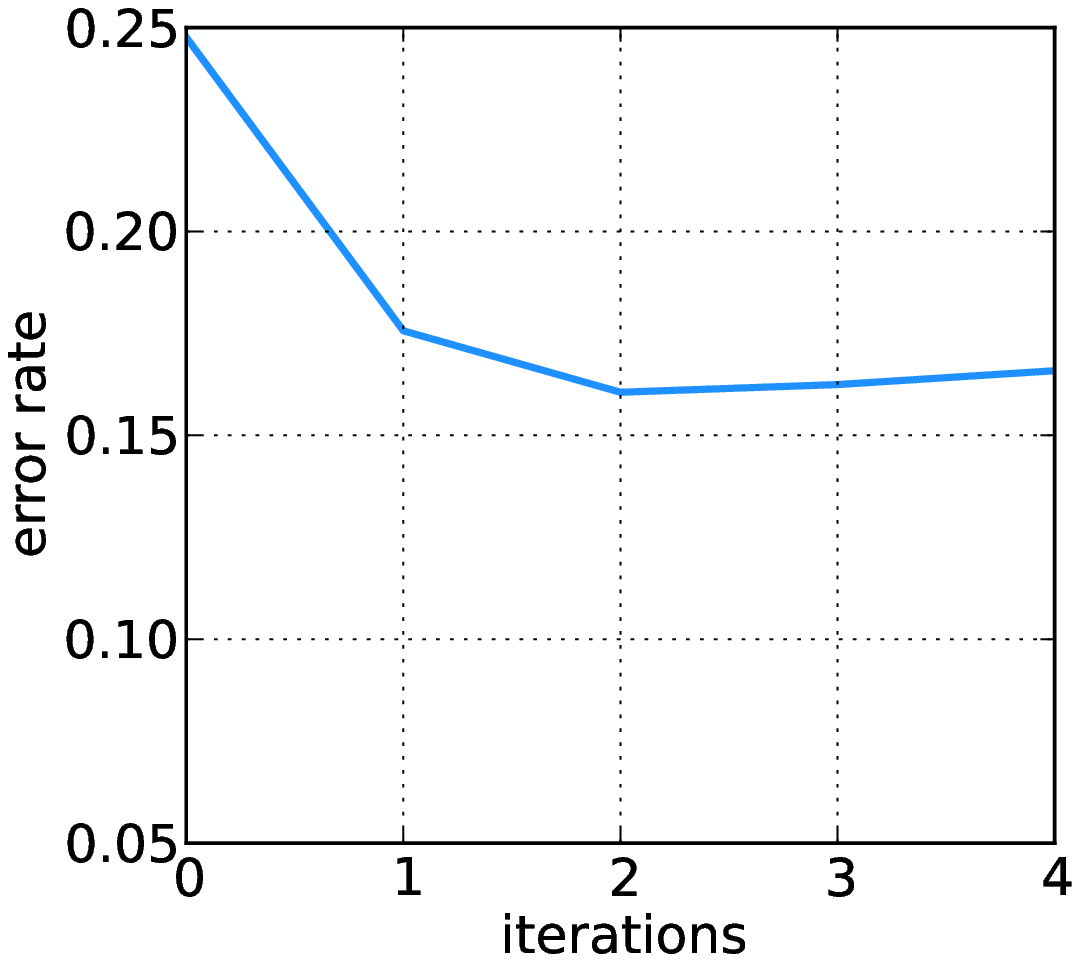}
}
\subfigure[RTE]{
\label{iter:c}
\includegraphics[scale=0.35]{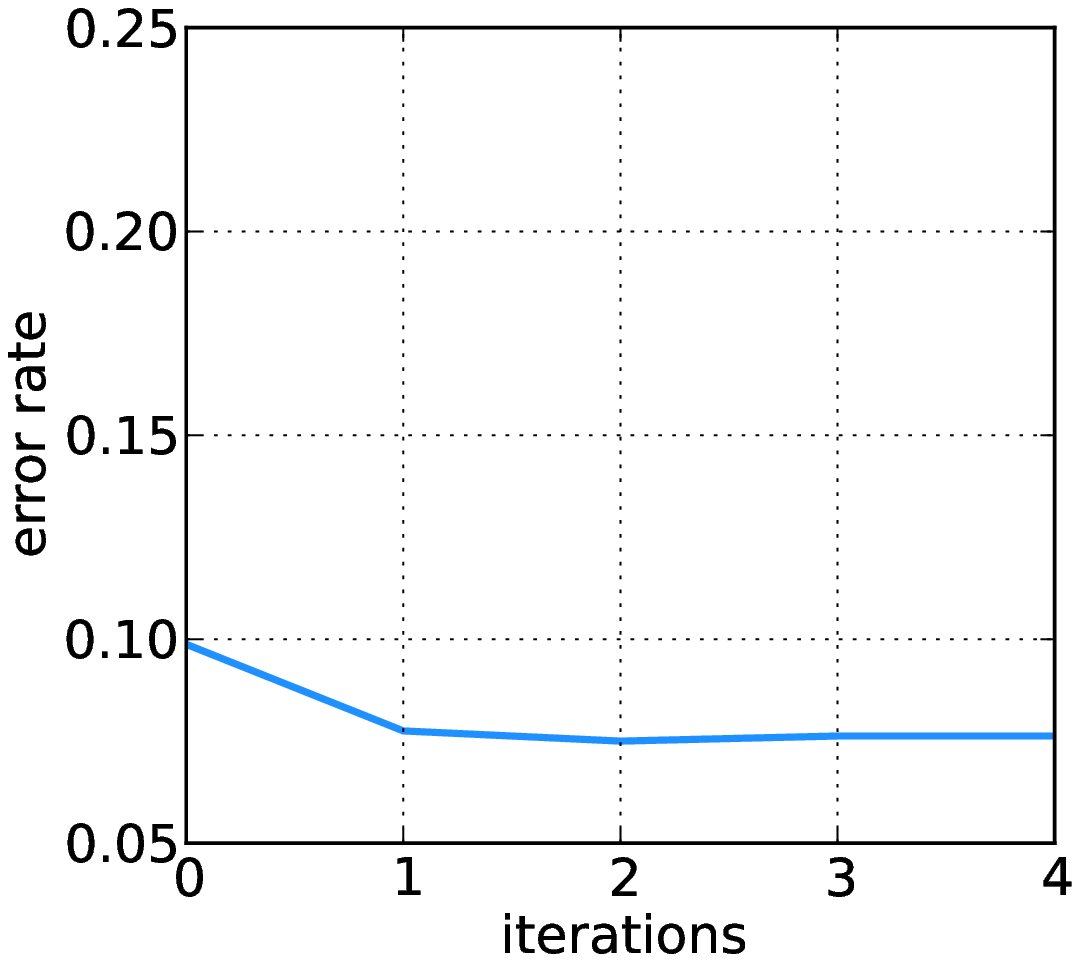}
}
\subfigure[Synthetic]{
\label{iter:d}
\includegraphics[scale=0.34]{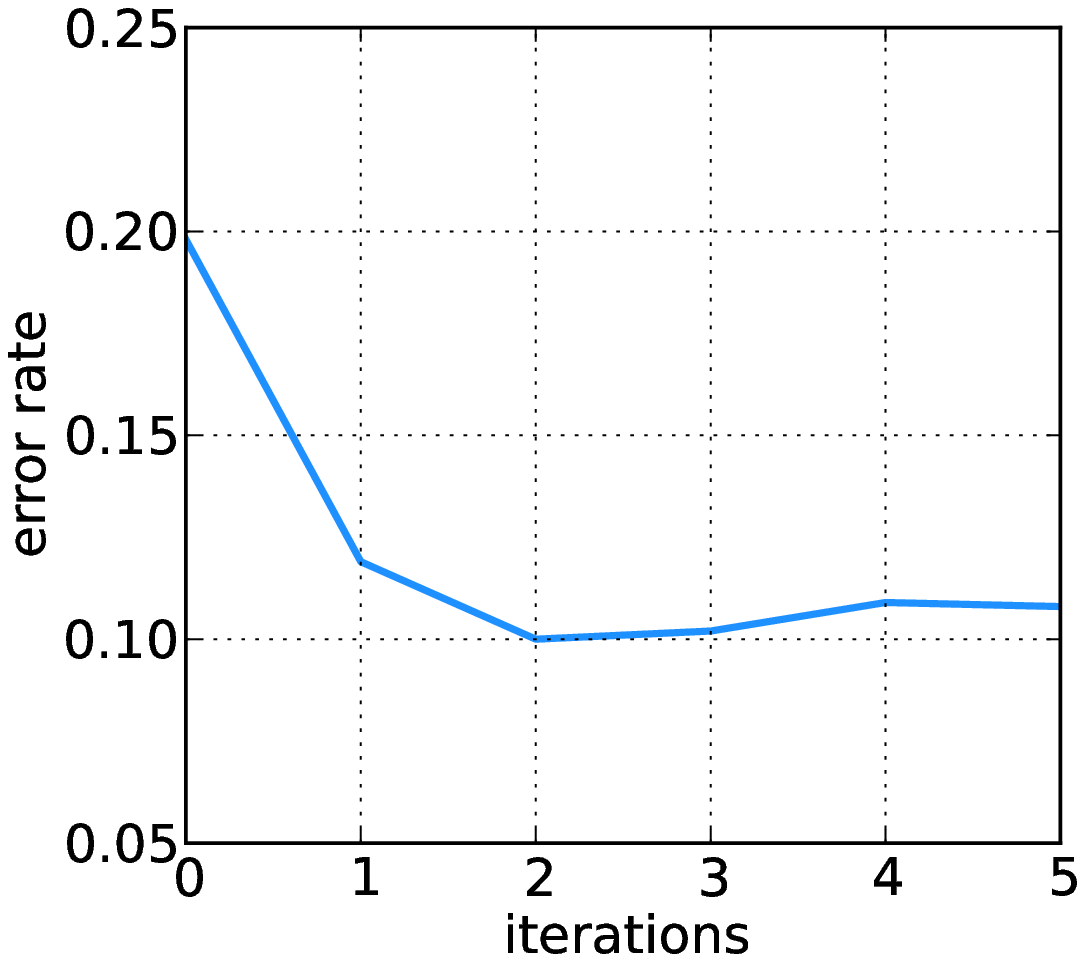}
}
\caption{Collapsed variational inference iteration process}
\label{cviter}
\end{figure}

In Section 4.3, we derive a collapsed variational inference algorithm for the IDBLA model. Now, we show the iteration processes of the algorithm in Figure~\ref{cviter}. We can see that the algorithm converges after only a few iterations. Note that the collapsed variation inference attains quite similar results with Gibbs sampling in our experiments.

\subsubsection{Negative Log Likelihood}

\begin{figure}[h]
\centering
\subfigure[Heartdisease]{
\label{NLLs:a}
\includegraphics[scale=0.3]{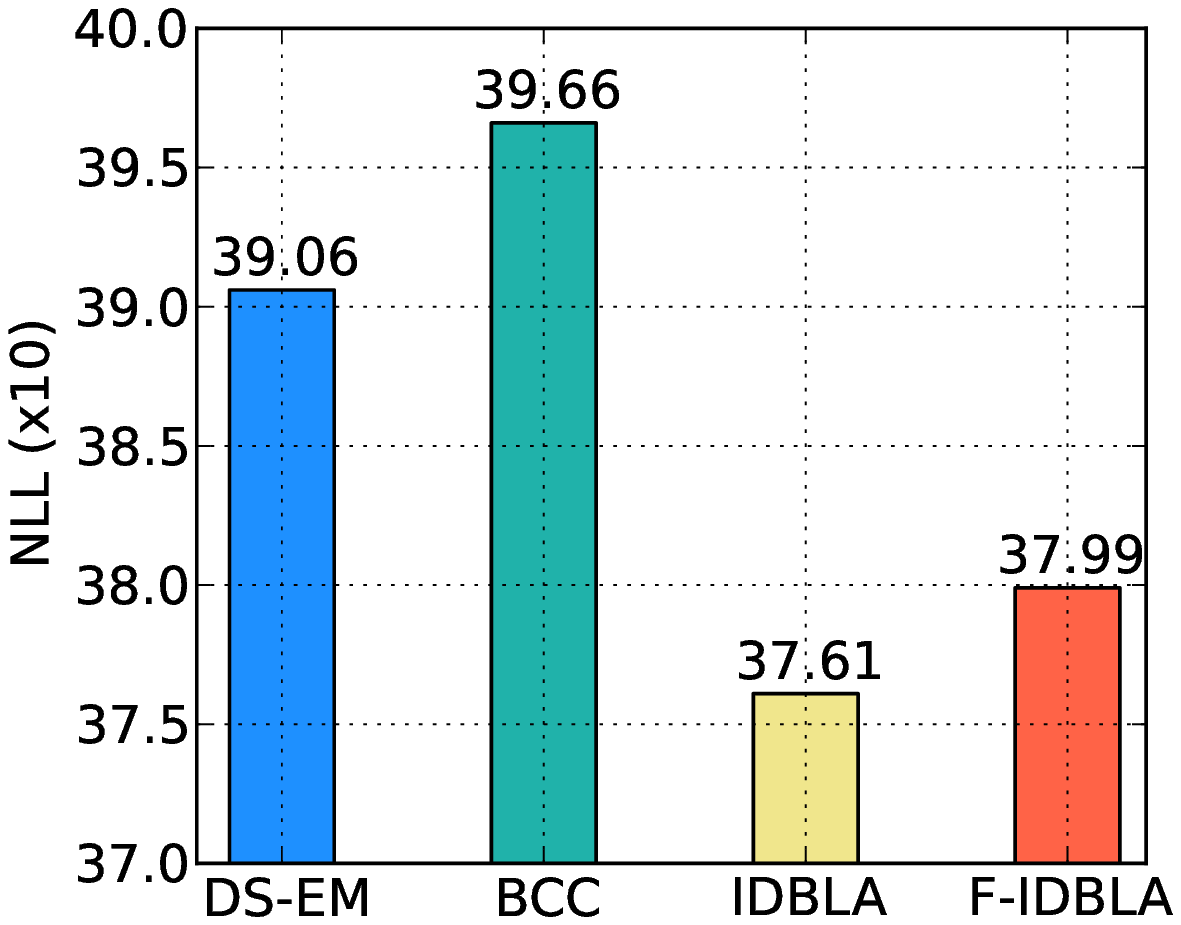}
}
\subfigure[Synthetic]{
\label{NLLs:b}
\includegraphics[scale=0.3]{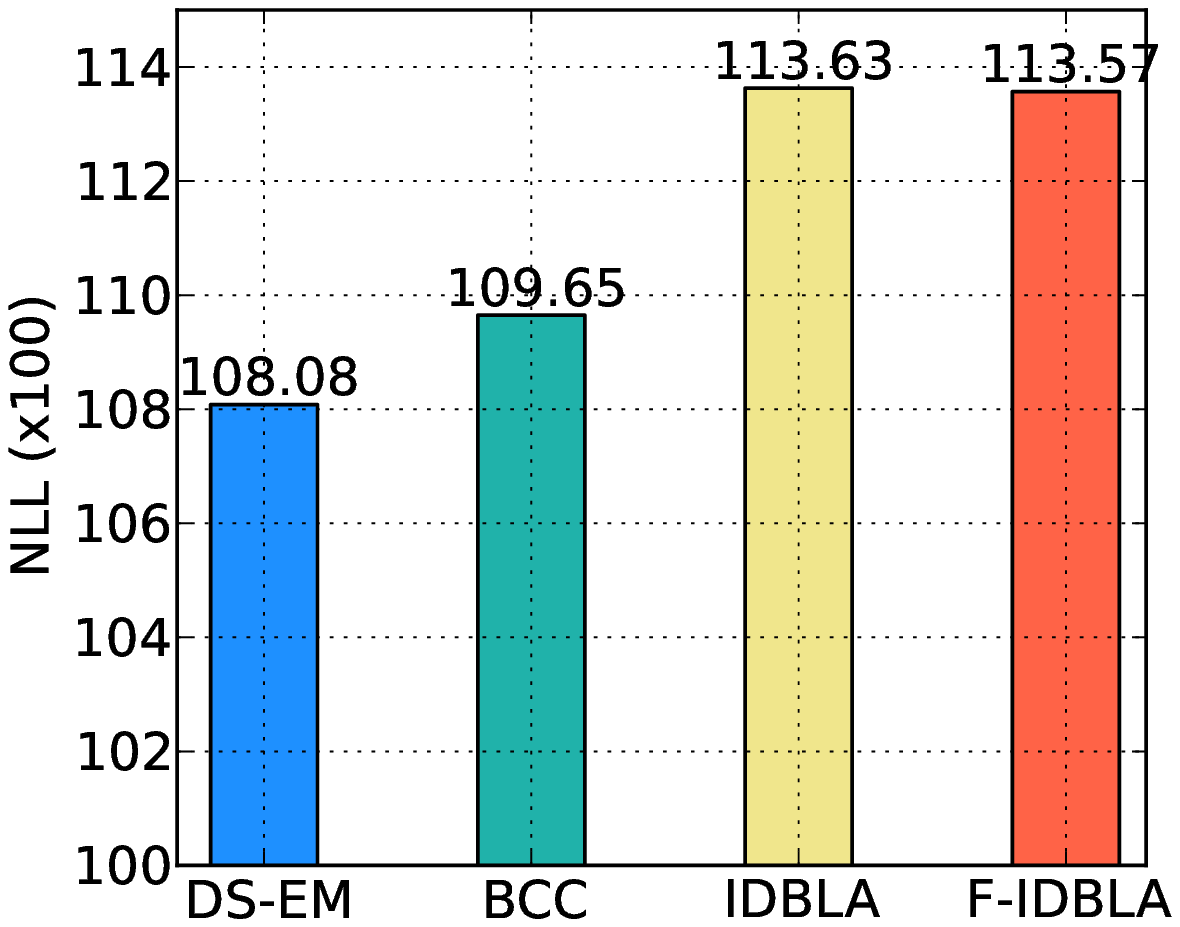}
}
\caption{NLLs of methods}
\label{NLLs}
\end{figure}

Then, we use negative log likelihood (NLL) to evaluate the methods. It provides a more comprehensive measure. NLL can simultaneously evaluate the true labels and confusion matrices found by the model. The likelihoods of IDBLA and Fixed-IDBLA are computed with:
\begin{equation*}
p(\bm{L}|\bm{\pi},\bm{T},\bm{Q})=\prod_{k=1}^{K}\prod_{h=1}^{H}\prod_{t=1}^{C}\prod_{c=1}^{C}(\pi_{t,c}^{(k,h)})^{\mathcal{N}_{l}(k,h,t,c)},
\end{equation*}
The likelihoods of DS-EM and BCC are computed with:
\begin{equation*}
p(\bm{L}|\bm{\phi},\bm{T}) = \prod_{i=1}^{I}\prod_{k=1}^{K}\prod_{l=1}^{C}(\phi_{T_i,l}^{(k)})^{\mathbbm{I}(L_{i,k}=l)}
\end{equation*}
where $\bm{\phi}^{(k)}$ is the confusion matrix for worker $k$. 
The NLLs of DS-EM, BCC and our methods on the Heartdisease dataset are shown in Figure~\ref{NLLs:a}. Figure~\ref{NLLs:b} shows the NLLs of the methods on the Synthetic dataset. The NLLs of the four methods are close to each other which means that our methods can also find high quality confusion matrices to evaluate the reliabilities and potential biases of workers.

\subsubsection{Model Analysis}

Finally, we further investigate the effectiveness of the parameter initialization and the quality of the difficulty level prediction of our models.

\textbf{Parameter Initialization Effectiveness}: 
We introduce the parameter initialization method in Section 4.4. In order to prove the effectiveness of the initialization, we remove it from our methods and then conduct the experiments again. For simplicity, we show the results on the Heartdisease dataset in Table~\ref{init}. For IDBLA model, the error rate increases from $0.160$ to $0.409$. For Fixed-IDBLA model, the error rate increases from $0.156$ to $0.430$.

\begin{table}[h]
\caption{Error-rates of Methods on the Heartdisease Dataset}
\label{init}
\begin{center}
\begin{tabular}{c|c|c}\hline\hline
Model & Initialization & Without Initialization \\ \hline
IDBLA & 0.160 & 0.409 \\
F-IDBLA & 0.156 & 0.430 \\ \hline
\end{tabular}
\end{center}
\end{table}

\textbf{Difficulty Level Prediction Quality}: Compare to the ground truth labels, we can get the labeling error rate $E_i$ of a item $i$. 
\begin{equation*}
E_i = \frac{\sum_{k=1}^{K}\mathbbm{I}(L_{i,k} \neq None, L_{i,k} \neq TRUTH_i) }{\sum_{k=1}^{K}\mathbbm{I}(L_{i,k} \neq None)}
\end{equation*}
where $TRUTH_i$ is the ground truth labels of item $i$. Average the labeling error rates of items in the same difficulty level, we can get the average labeling error rate of the difficulty level. 
We illustrate the results on the Heartdisease dataset. For IDBLA model, the average labeling error rate of the predicted easiest difficulty level is 0.280 and the average labeling error rate of the predicted hardest difficulty level is 0.523. For Fixed-IDBLA model, the average labeling error rate of the predicted easiest difficulty level is 0.213, the average labeling error rate of the predicted hardest difficulty level is 0.694. All the prediction results are consistent with the truth. That means the prediction of item difficulty is effective.

\section{Conclusions}

We propose the IDBLA model to aggregate labels collected from non-professional workers. In this model, the item difficulties are taken into consideration. Each worker-difficulty level pair is associated with a confusion matrix. The model finds the values of the confusion matrices and other latent variables in the learning process, and further uses them to infer the true labels. We derive a collapsed variational inference algorithm for the IDBLA model. The algorithm converges within only a few iterations and attains good solutions. We define a variation of the IDBLA model which assumes that there exists some very easy items and some very difficult items. We also design a method to preliminarily predict the true label and difficulty of each item. The prediction results are used to initialize the latent parameters. The experiments are conducted on three real datasets and one synthetic dataset. The empirical results show that our methods are effective.


\appendix
\section{Appendix}
Actually, we derive the equations (17) and (18) according to the variational Bayes algorithm \cite{bishop2006pattern}. To be self-contained we show the derivation of equation (18). Note that the derivation of equation (17) is similar. According to equation (13) we have:
\begin{multline}
\mathcal{L}(q(\bm{T},\bm{Q})) = \int q(\bm{T},\bm{Q})\log \frac{ p(\bm{T},\bm{Q},\bm{L}|\bm{\mu})}{q(\bm{T},\bm{Q})} d_{\bm{T},\bm{Q}} = \\
\int \Big \{ \prod_{j=1}^{I}q(T_j|\bm{\lambda}_j)q(Q_j|\bm{\rho}_j) \Big \}
\Big \{ \log p(\bm{T},\bm{Q},\bm{L}|\bm{\mu}) - \\
\sum_{j=1}^{I} \big ( \log q(T_j|\bm{\lambda}_j) + \log q(Q_j|\bm{\rho}_j) \big ) \Big \}  
 d_{\bm{T},\bm{Q}} = \\
 \int q(Q_i|\bm{\rho}_i) \Big \{ \int \log p(\bm{T},\bm{Q},\bm{L}|\bm{\mu}) \big ( \prod_{j=1}^{I}q(T_j|\bm{\lambda}_j) d_{T_j} \big ) \big ( \prod_{j \neq i}q(Q_j|\bm{\rho}_j) d_{Q_j} \big ) \Big \} d_{Q_i} \\
 - \int q(Q_i|\bm{\rho}_i) \log q(Q_i|\bm{\rho}_i)d_{Q_i} + const
\end{multline}
where
$\int \log p(\bm{T},\bm{Q},\bm{L}|\bm{\mu}) \big ( \prod_{j=1}^{I}q(T_j|\bm{\lambda}_j) d_{T_j} \big ) \big ( \prod_{j \neq i}q(Q_j|\bm{\rho}_j) d_{Q_j} \big ) = \\ \mathbbm{E}_{q(\bm{T},\bm{Q}^{\lnot i})}[\log p(\bm{T},\bm{Q},\bm{L}|\bm{\mu})]$.

Defining a new distribution $\tilde{p}(\bm{L},Q_i)$ by the relation $\log \tilde{p}(\bm{L},Q_i) = \mathbbm{E}_{q(\bm{T},\bm{Q}^{\lnot i})}[\log p(\bm{T},\bm{Q},\bm{L}|\bm{\mu})] + const$. Then 
$$
\mathcal{L}(q(\bm{T},\bm{Q})) = - KL(q(Q_i|\bm{\rho}_i)|\tilde{p}(\bm{L},Q_i)) + const.
$$
We maximize $\mathcal{L}(q(\bm{T},\bm{Q}))$ with respect to $q(Q_i|\bm{\rho}_i)$. The maximum is achieved at $q(Q_i|\bm{\rho}_i)=\tilde{p}(\bm{L},Q_i)$, then we have:
\begin{multline*}
q(Q_i|\bm{\rho}_i) \propto \exp (\mathbbm{E}_{q(\bm{T},\bm{Q}^{\lnot i})}[\log p(\bm{T},\bm{Q},\bm{L}|\bm{\mu})]) \\
= \exp (\mathbbm{E}_{q(\bm{T},\bm{Q}^{\lnot i})}[\log p(Q_i|rest)]+\mathbbm{E}_{q(\bm{T},\bm{Q}^{\lnot i})}[\log p(rest)])\\
\propto \exp(\mathbbm{E}_{q(\bm{T},\bm{Q}^{\lnot i})}[\log p(Q_i|rest)])
\end{multline*}
where $rest$ means $\bm{T}, \bm{Q}^{\lnot i}, \bm{L}$ and $\bm{\mu}$. Plugging in $Q_i = h$, we have equation (18).
\begin{acks}
~

~

~

~

~

~

~

~

~
\end{acks}

\bibliographystyle{ACM-Reference-Format}

\end{document}